\documentclass{bmvc2k}

\title{Triangulation: Why Optimize?}

\addauthor{Seong Hun Lee}{seonghunlee@unizar.es}{1}
\addauthor{Javier Civera}{jcivera@unizar.es}{1}

\addinstitution{
 I3A\\
 University of Zaragoza\\
 Zaragoza, Spain\\
}

\runninghead{Lee, Civera}{Triangulation: Why Optimize?}

\usepackage{lipsum}
\usepackage{wrapfig,graphicx}
\usepackage{bm}
\let\mathcal\undefined
\DeclareMathAlphabet{\mathcal}{OMS}{cmsy}{m}{n}
\usepackage{amssymb}
\usepackage{stackrel}

\usepackage{enumitem}
\usepackage{tabularx, booktabs}
\usepackage{multirow}
\newcolumntype{Y}{>{\centering\arraybackslash}X}

\usepackage{ntheorem}
\theoremstyle{break}

\newtheorem{lemma}{Lemma}
\newcommand*{\QEDA}{\hfill\ensuremath{\blacksquare}}%
\begin{document}
\maketitle
\vspace{-1em}
\begin{abstract}
\vspace{-0.5em}
For decades, it has been widely accepted that the gold standard for two-view triangulation is to minimize the cost based on reprojection errors.
In this work, we challenge this idea.
We propose a novel alternative to the classic midpoint method that leads to significantly lower 2D errors and parallax errors.
It provides a numerically stable closed-form solution based solely on a pair of backprojected rays. Since our solution is rotationally invariant, it can also be applied for fisheye and omnidirectional cameras.
We show that for small parallax angles, our method outperforms the state-of-the-art in terms of combined 2D, 3D and parallax accuracy, while achieving comparable speed.
\end{abstract}
\vspace{-2em}
%-------------------------------------------------------------------------
\section{Introduction}
\label{sec:introduction}
\vspace{-0.5em}
Locating the 3D point given its projections in multiple views is called triangulation.
This classic yet fundamental problem in computer vision has immediate relevance to many applications, including visual odometry \cite{svo}, simultaneous localization and mapping (SLAM) \cite{orb-slam} and structure-from-motion (SfM) \cite{sfm_revisited}.
As such, achieving fast and accurate triangulation has been a goal of many research endeavors in the past decades.

For two views of known calibration and pose, the problem 
could be solved ideally if one finds the intersection of two backprojected rays corresponding to the same point.
However, the two rays are most likely skew due to noisy measurements and inaccurate camera model.
Since it is not obvious how to estimate the 3D position of the point from two skew rays, different methods have been proposed.
Mainly, they can be classified into three types:
(1) midpoint methods \cite{beardsley_midpoint1, beardsley_midpoint2, iteratively_reweighted_midpoint} that find the (weighted) midpoint of the common perpendicular between the two rays, (2) linear least squares methods \cite{hartley_triangulation}, and (3) optimal methods that ``minimally'' correct the two rays to make them intersect \cite{kanatani,hartley_triangulation, closed_form_optimal_triangulation_based_angular_errors}.
Note that all these three types of methods produce solutions that minimize some cost function;
the (weighted) midpoint minimizes the (weighted) sum of squared distances to each ray, linear least squares methods minimize the algebraic errors, and optimal methods minimize a cost function based on either \textit{image} reprojection errors \cite{kanatani, hartley_triangulation, lindstrom} or \textit{angular} reprojection errors \cite{closed_form_oliensis, closed_form_optimal_triangulation_based_angular_errors}.
The most common cost functions are the $L_1$ norm (sum of magnitude), $L_2$ norm (sum of squares) and $L_\infty$ norm (maximum) of the reprojection errors.
%Today, it is widely believed that triangulation from two views is well established and understood.
%As a result, the focus of recent research endeavors has shifted towards multiview triangulation \cite{L2_kang, L_inf_dai, iteratively_reweighted_midpoint} and bundle adjustment \cite{ba_modern_synthesis, ba_revisited}.

In this work, we suggest a different approach.
Instead of minimizing geometric or algebraic errors, we find a midpoint between a certain pair of points on each ray.
Like the classic midpoint method, our method takes the two rays as input.
Therefore, it is invariant to changes of camera rotation and applicable for perspective, fisheye and omnidirectional cameras.
Unlike the classic method, however, the two points on each ray are not necessarily on the common perpendicular.
We will see that our midpoint method bears a striking similarity to the classic method in the formulation, and yet it offers a significant performance gain in 2D and parallax accuracy. 
Compared to the optimal methods, our method yields much lower 3D errors at low parallax and similar 2D errors to those of $L_2$ and $L_\infty$ optimal methods.
This motivates the question: 
In two-view triangulation, why optimize if there is a better way?

The main contributions of this paper are the following:
\vspace{-0.5em}
\begin{itemize}\itemsep0em
    \item We propose a novel method belonging to a group called the generalized weighted midpoint (GWM) method.
    We show that our method outperforms existing ones (including the classic midpoint method and the state-of-the-art optimal methods) in terms of combined 2D, 3D and parallax accuracy. 
    \vspace{-0.2em}
    \item Additionally, we propose a test of the adequacy (similar to the cheirality check) that identifies unreliable results and a weighting scheme that enhances 2D accuracy. 
    \vspace{-0.2em}
    \item We perform an extensive evaluation and analysis of various methods, revealing an intricate link between 3D accuracy and parallax estimation.
    This will provide an intuitive explanation of why our midpoint method performs better than the others.
\end{itemize}
\vspace{-1.5em}

% %% Paper Organization
% The rest of the paper is organized as follows:
% In Section \ref{sec:related_work}, we review the related work.
% Section \ref{sec:preliminaries} introduces the notation and preliminaries. 
% We describe and evaluate the proposed method in Section \ref{sec:method} and \ref{sec:results}, respectively.
% Finally, the conclusions are given in Section \ref{sec:conclusion}. 

\section{Related Work}
\label{sec:related_work}
\vspace{-0.5em}
One of the earliest works that addressed the two-view triangulation problem is \cite{eight_point} where the depth of a 3D point is estimated using simple algebra.
For robustness, later works mostly adopted geometric approaches, such as the midpoint method \cite{beardsley_midpoint1, beardsley_midpoint2} and the minimization of the epipolar distance \cite{harris1, harris2} or reprojection error \cite{hartley_euclidean}.
Among those, the last approach has become the \textit{de facto} standard in computer vision \cite{hartley_book}.

\textit{Optimal methods} refer to those triangulation methods that minimize the cost based on reprojection errors.  
Assuming that the image measurements are independently perturbed by the noise in the same distribution of certain types, the optimal methods find the maximum likelihood (ML) solution.
For Gaussian and Laplacian distribution, the ML solution is to minimize the $L_2$ norm or $L_1$ norm of the reprojection errors, respectively \cite{L1_mle}.
This can be found in closed form by solving a polynomial of degree six or eight \cite{hartley_triangulation}.
Alternatively, the $L_2$ solution can be obtained using iterative correction methods \cite{kanatani, lindstrom}.
While these iterative methods do not guarantee global optimality, they were shown to be faster and more stable.
For a uniform distribution, minimizing the $L_\infty$ norm leads to the ML estimate for the lower bound of the noise \cite{optimal_algorithms_hartley}, and the solution is obtained by solving a quartic polynomial \cite{nister_phd_thesis}.
Unlike the $L_1$ and $L_2$ cost, the $L_\infty$ cost has a simple shape with a single minimum, but it is relatively more sensitive to noise and outliers \cite{L_inf_hartley}.
%All these methods are affine and projective invariant \cite{hartley_triangulation, nister_phd_thesis}.

These optimal methods assume that the image measurement errors follow certain distributions. 
However, this assumption is neither justified nor likely \cite{optimal_algorithms_hartley}.
An equally (if not more) justified alternative is to assume that the noise model applies to the bearing measurements instead of the image.
For fisheye or omnidirectional cameras, the angular reprojection errors are more suitable than the image reprojection errors \cite{closed_form_oliensis, sfm_wideangle}.
Also, formulating the triangulation problem in terms of angular errors leads to much simpler ML solutions \cite{closed_form_oliensis, closed_form_optimal_triangulation_based_angular_errors}.

Although the existing optimal methods can provide relatively good 3D results in many cases \cite{hartley_triangulation, closed_form_optimal_triangulation_based_angular_errors}, none of them are theoretically optimal in terms of 3D errors.
In fact, the discrepancy between 2D optimality and 3D accuracy has already been reported by Hartley and Sturm \cite{hartley_triangulation}.
They found that in Euclidean reconstruction, the midpoint and the linear least squares method achieve higher 3D accuracy than the $L_1$ or $L_2$ optimal methods, despite consistently (and sometimes significantly) larger 2D errors. 
In this work, we provide additional insights on this matter.
Furthermore, we show that a simple modification to the midpoint method can substantially reduce the 2D errors while maintaining 3D accuracy.

\newpage
\section{Preliminaries}
\label{sec:preliminaries}

\begin{figure}[t]
 \centering
 \includegraphics[width=\textwidth]{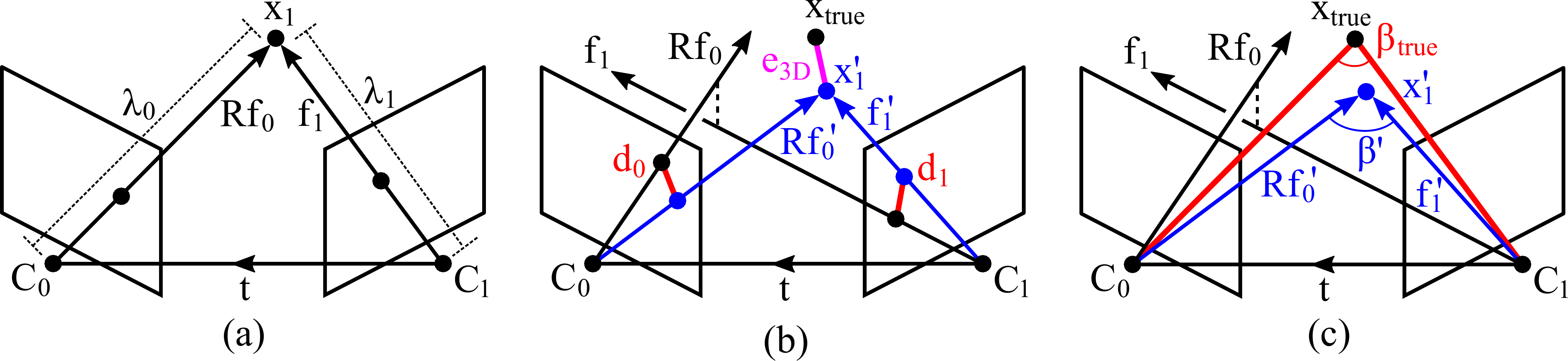}
\caption{
\textbf{(a)} Epipolar geometry when the two backprojected rays intersect.
All vectors are in the coordinate system of $C_1$.
\textbf{(b)} 3D error ($e_\text{3D}$) measures the Euclidean distance between the estimate ($\mathbf{x}'_1$) and the true position of the point ($\mathbf{x}_\text{true}$), while 2D error ($d_0$, $d_1$) measures the offset between the observation and the reprojection of the estimated point in each frame.
\textbf{(c)} After the triangulation, one can estimate the parallax ($\beta'$) from the corrected rays.
}
\label{fig:epipolar_geometry}
\vspace{-1em}
\end{figure}

\vspace{-0.5em}
Throughout the paper, we use bold letters for vectors and matrices, and light letters for scalars.
The Euclidean norm of a vector $\mathbf{v}$ is denoted by $\lVert \mathbf{v} \rVert$, and the unit vector by $\widehat{\mathbf{v}}={\mathbf{v}}/{\lVert\mathbf{v}\rVert}$.
The angle between two lines $\mathbf{L}_0$ and $\mathbf{L}_1$ is denoted by $\angle\left(\mathbf{L}_0, \ \mathbf{L}_1\right)\in[0, \pi/2]$.

Consider a 3D point observed by two cameras $C_0$ and $C_1$.
We define $\mathbf{x}_0 = [x_0, y_0, z_0]^\intercal$ and $\mathbf{x}_1 = [x_1, y_1, z_1]^\intercal$ as the unknown 3D coordinates of the point in the camera reference frame $C_0$ and $C_1$, respectively. 
Let $\mathbf{R}$ and $\mathbf{t}$ be the known rotation and translation between the two cameras, such that $\mathbf{x}_1=\mathbf{R}\mathbf{x}_0+\mathbf{t}$.
Assuming that the camera calibration matrix $\mathbf{K}$ is known, the normalized image coordinates $\mathbf{f}_0=[x_0/z_0, y_0/z_0, 1]^\intercal$ and $\mathbf{f}_1=[x_1/z_1, y_1/z_1, 1]^\intercal$ can be obtained by $\mathbf{f}_0 = \mathbf{K}^{-1}\mathbf{u}_0$ and $\mathbf{f}_1 = \mathbf{K}^{-1}\mathbf{u}_1$, where $\mathbf{u}_0 = (u_0, v_0,1)^\intercal$ and $\mathbf{u}_1=(u_1, v_1,1)^\intercal$ are the homogeneous pixel coordinates of the point observation in each frame.

In the ideal situation (Fig. \ref{fig:epipolar_geometry}{\color{red}a}), the two backprojected rays intersect, satisfying the epipolar constraint \cite{eight_point}, i.e.,
${\mathbf{f}_1}\cdot\left(\mathbf{t}\times\mathbf{R}\mathbf{f}_0\right)=0$.
Then, the intersection is given by $\mathbf{x}_1 = \lambda_0\mathbf{R}\widehat{\mathbf{f}}_0+\mathbf{t}$ or $\mathbf{x}_1 = \lambda_1\widehat{\mathbf{f}}_1$ for some scalar depth $\lambda_0$ and $\lambda_1$.
However, this rarely happens due to inaccuracies in the image measurements and the camera model.
Inferring a 3D point from two skew rays requires a nontrivial method.

Once the estimate of the 3D point ($\mathbf{x}'_1$) is obtained using some triangulation method, its accuracy can be evaluated in several ways.
One way is to compute the 3D error, i.e., $e_{3D} = \lVert \mathbf{x}'_1-\mathbf{x}_\text{true}\rVert$.
Another way is to compute the 2D error (aka the reprojection error), i.e.,
\vspace{-0.5em}
\begin{equation}
    \label{eq:reprojection_error}
    d_i 
    = \lVert\mathbf{K}\left(\mathbf{f}_i-{\mathbf{f}_i}'\right)\rVert
    =\left\Vert\mathbf{K}\left(\mathbf{f}_i-\left([0 \ 0 \ 1] \ \mathbf{x}'_i\right)^{-1}\mathbf{x}'_i\right)\right\Vert \quad \text{for} \quad i=0,1,
    \vspace{-0.5em}
\end{equation}
where $\mathbf{x}'_0 = \mathbf{R}^\intercal\left(\mathbf{x}'_1-\mathbf{t}\right)$.
These two errors are illustrated in Fig. \ref{fig:epipolar_geometry}{\color{red}b}.
Note that the 2D error represents the deviation from the measurement, whereas the 3D error represents the deviation from the ground truth.
Also, unlike the 3D error, the 2D error of a 3D point can be evaluated in different norms, e.g., $L_1$ norm $\left(d_0+d_1\right)$, $L_2$ norm $\left(\sqrt{ d_0^2+d_1^2}\right)$ and $L_\infty$ norm $\left(\max(d_0, d_1)\right)$.
Besides 2D and 3D accuracy, we can also evaluate the accuracy of the resulting parallax angle (see Fig. \ref{fig:epipolar_geometry}{\color{red}c}). 
The parallax error is defined as follows:
\vspace{-0.5em}
\begin{equation}
    e_\beta 
    = |\beta_\text{true}-\beta'|
    = |\angle\left(\mathbf{x}_\text{true}, \mathbf{x}_\text{true}-\mathbf{t}\right)-\angle\left(\mathbf{x}'_1,\mathbf{x}'_1-\mathbf{t}\right)|.
    \vspace{-0.5em}
\end{equation}
We define the ``raw parallax'' as the angle between the original backprojected rays:
\vspace{-0.5em}
\begin{equation}
\label{eq:raw_parallax}
    \beta_\text{raw}=\angle\left(\mathbf{Rf}_0, \mathbf{f}_1\right).
\vspace{-0.5em}
\end{equation}
This gives a rough estimate of the parallax angle independently of the translation and the triangulation method.

\newpage
\section{Proposed method}
\label{sec:method}
\subsection{Generalized Weighted Midpoint (GWM) Method}
\label{subsec:generalized_weighted_midpoint_method}

\begin{wrapfigure}{R}{0.5\textwidth}
 \centering
 \vspace{-1em}
 \includegraphics[width=0.5\textwidth]{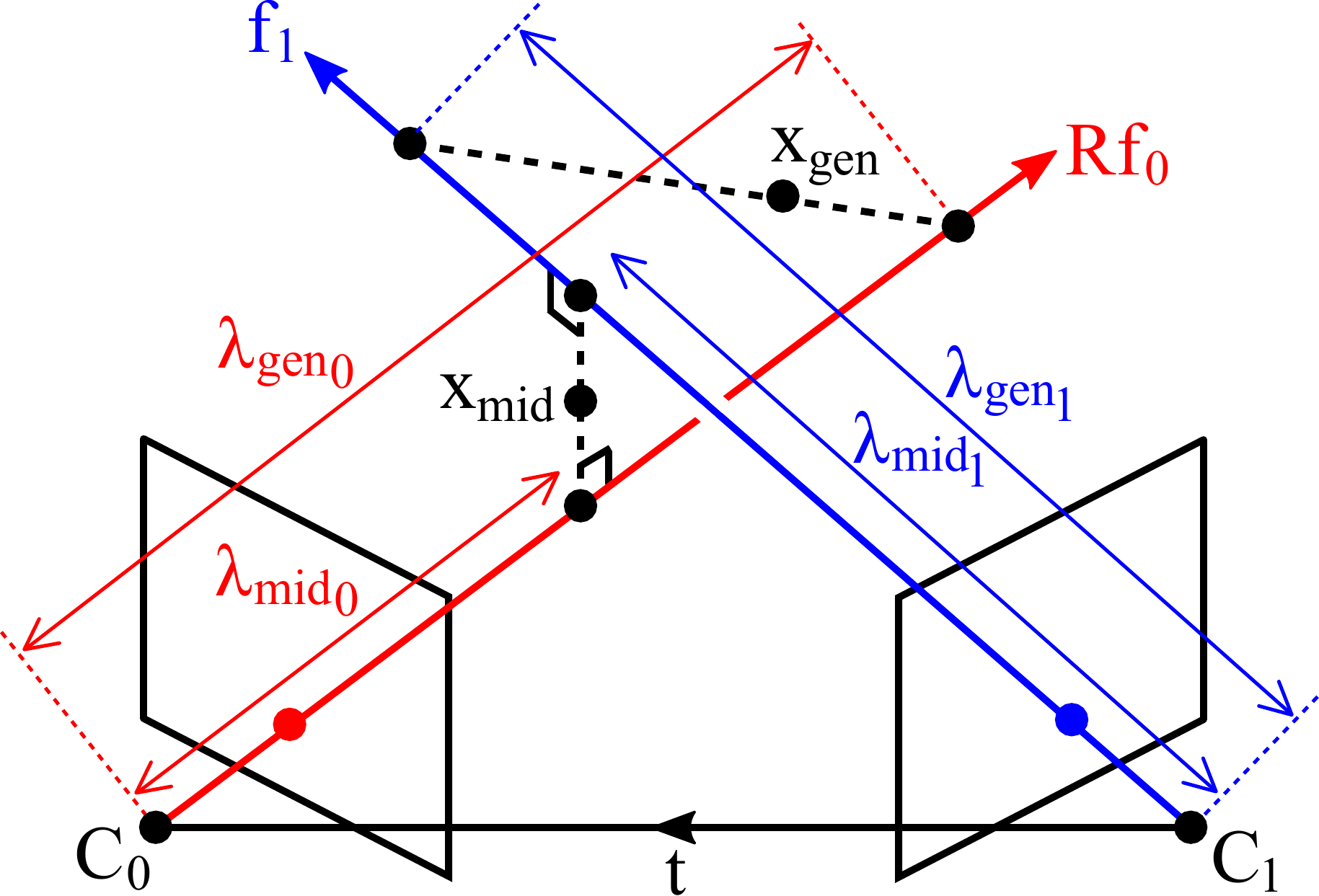}
\caption{The classic midpoint and another example of the generalized weighted midpoint.}
\vspace{-1em}
\label{fig:generalized_weighted_midpoint}
\end{wrapfigure}

A GWM method consists of three steps:
(1) Given two backprojected rays corresponding to the same point, estimate the depth along each ray ($\lambda_0$, $\lambda_1$) using some method.
(2) Compute the 3D point on each ray at depth $\lambda_0$ and $\lambda_1$, i.e., $\mathbf{t}+\lambda_0\mathbf{R}\widehat{\mathbf{f}}_0$ and $\lambda_1\widehat{\mathbf{f}}_1$ in $C_1$.
(3) Obtain the final estimate of the  3D point by computing their weighted average. 

The classic midpoint method \cite{beardsley_midpoint1, beardsley_midpoint2, hartley_triangulation} is one such type of method where the two points on each ray are the closest pair of points with the equal weight.
Fig. \ref{fig:generalized_weighted_midpoint} shows another possible example of the generalized weighted midpoint.

\subsection{Alternative Midpoint Method}
\label{subsec:alternative_midpoint_method}

We propose an alternative midpoint method that belongs to the generalized weighted midpoint method.
First, consider the case where the two backprojected rays happen to intersect (see Fig. \ref{fig:epipolar_geometry}{\color{red}a}).
In this case, the most sensible solution is the point of intersection, and the corresponding depths along the rays can be obtained using the sine rule:   
\begin{equation}
\label{eq:lambdas1}
    \lambda_0 = 
    \frac{\sin{\left(\angle\left( \mathbf{f}_1, \mathbf{t}\right)\right)}}{\sin{\left(\angle\left(\mathbf{Rf}_0, \mathbf{f}_1,\right)\right)}}\lVert\mathbf{t}\rVert=\frac{\lVert\widehat{\mathbf{f}}_1\times\mathbf{t}\rVert}{\lVert\mathbf{R}\widehat{\mathbf{f}}_0\times\widehat{\mathbf{f}}_1\rVert}, \quad
    \lambda_1 = 
    \frac{\sin{\left(\angle\left( \mathbf{Rf}_0, \mathbf{t}\right)\right)}}{\sin{\left(\angle\left(\mathbf{Rf}_0, \mathbf{f}_1,\right)\right)}}\lVert\mathbf{t}\rVert=\frac{\lVert\mathbf{R}\widehat{\mathbf{f}}_0\times\mathbf{t}\rVert}{\lVert\mathbf{R}\widehat{\mathbf{f}}_0\times\widehat{\mathbf{f}}_1\rVert}.
\end{equation}
We use this formula to estimate the depths even when the two rays are skew.
Computing the 3D points on each ray at depth $\lambda_0$ and $\lambda_1$, respectively, we get
\begin{equation}
    \mathbf{t}+\lambda_0\mathbf{R}\widehat{\mathbf{f}}_0 = \mathbf{t}+ \frac{\lVert{\mathbf{f}}_1\times\mathbf{t}\rVert}{\lVert\mathbf{R}{\mathbf{f}}_0\times{\mathbf{f}}_1\rVert}\mathbf{R}{\mathbf{f}}_0 \quad \text{and} \quad
    \lambda_1\widehat{\mathbf{f}}_1 = \frac{\lVert{\mathbf{Rf}}_0\times\mathbf{t}\rVert}{\lVert\mathbf{R}{\mathbf{f}}_0\times{\mathbf{f}}_1\rVert}{\mathbf{f}}_1,
\end{equation}
Taking the midpoint between these two points leads to
\begin{equation}
\label{eq:unweighted_midpoint}
    \mathbf{x}'_1 = \frac{1}{2}\left(\mathbf{t}+ \frac{\lVert{\mathbf{f}}_1\times\mathbf{t}\rVert}{\lVert\mathbf{R}{\mathbf{f}}_0\times{\mathbf{f}}_1\rVert}\mathbf{R}{\mathbf{f}}_0+\frac{\lVert{\mathbf{Rf}}_0\times\mathbf{t}\rVert}{\lVert\mathbf{R}{\mathbf{f}}_0\times{\mathbf{f}}_1\rVert}{\mathbf{f}}_1\right).
\end{equation}
Note that letting $\mathbf{p} = \mathbf{R}\widehat{\mathbf{f}}_0\times\widehat{\mathbf{f}}_1$, $\mathbf{q} = \mathbf{R}\widehat{\mathbf{f}}_0\times\mathbf{t}$ and $\mathbf{r} =  \widehat{\mathbf{f}}_1\times\mathbf{t}$ allows us to write \eqref{eq:lambdas1} as
\begin{equation}
\label{eq:lambdas2}
    \lambda_{0} = \frac{\lVert\mathbf{r}\rVert}{\lVert\mathbf{p}\rVert},
    \quad
    \lambda_{1} = \frac{\lVert\mathbf{q}\rVert}{\lVert\mathbf{p}\rVert}.
\end{equation}
Interestingly, these are in a similar form to the depths given by the classic midpoint method\footnote{We provide the derivation in the Appendix.}:
\begin{equation}
\label{eq:lambdas_mid}
    {\lambda_\text{mid}}_0 = \frac{\widehat{\mathbf{p}}\cdot\mathbf{r}}{\lVert\mathbf{p}\rVert},
    \quad
    {\lambda_\text{mid}}_1 = \frac{\widehat{\mathbf{p}}\cdot\mathbf{q}}{\lVert\mathbf{p}\rVert}.
\end{equation}
The difference is in the numerator; 
\eqref{eq:lambdas2} has the magnitude of $\mathbf{r}$ and $\mathbf{q}$, whereas \eqref{eq:lambdas_mid} has their projection onto $\mathbf{p}$.
As a result, we always get $\lambda_0 \geq {\lambda_\text{mid}}_0$ and $\lambda_1 \geq {\lambda_\text{mid}}_1$.
In most cases, this means that our midpoint will be located farther than the classic midpoint.
Fig. \ref{fig:generalized_weighted_midpoint} depicts one such example.
When we estimate that a point is located farther away from the cameras, it usually results in a lower estimate of the parallax angle, as will be shown in Section \ref{sec:results}.

\vspace{-0.5em}
\subsection{Cheirality and Test of Adequacy}
\label{subsec:cheirality}
\begin{figure}[t]
 \centering
 \includegraphics[width=\textwidth]{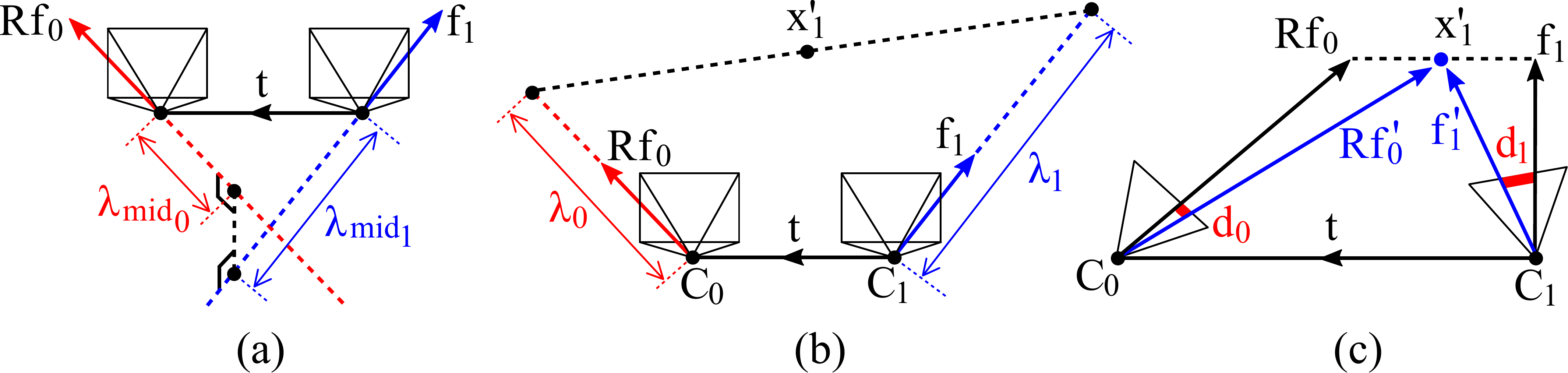}
\caption{
\textbf{(a)} A scenario where the classic midpoint method gives negative depths. 
The cheirality check will identify and remove this point. 
\textbf{(b)} In the same situation, our midpoint method will give positive depths.
The triangulation result satisfies the cheirality constraint, but it is most likely inaccurate.
\textbf{(c)} For unweighted midpoint methods, the frame with a smaller depth tends to get a larger reprojection error. 
In this example, $\lambda_1 < \lambda_0$ and $d_1 > d_0$.
}
\label{fig:cheirality_and_weighting}
\end{figure}

We say that the cheirality constraint \cite{hartley_book} is violated when a triangulated point has negative depth(s).
This can happen for many reasons, such as spurious data association or the noise in the image point near the epipole.
Normally, it does not pose a serious problem because we can easily check the cheirality for each point and discard the bad ones.
For the classic midpoint method, this can be done by checking the signs of the depths given by \eqref{eq:lambdas_mid}.
%\footnote{This applies for general camera models. For perspective cameras, one should also check whether the triangulated 3D point is within the field of view of both cameras.}.
For our midpoint method, however, this is not possible because the depths given by \eqref{eq:lambdas2} are always positive. 
Fig \ref{fig:cheirality_and_weighting}{\color{red}a} and  \ref{fig:cheirality_and_weighting}{\color{red}b} illustrate the difference between the two methods. 
In our method, the depths alone cannot tell us whether or not the triangulation result is reliable.

Therefore, we use a different method to test the adequacy;
we discard the point correspondence if changing the sign of at least one depth to negative leads to a smaller distance between the two points on each ray, i.e.,
\vspace{-0.2em}
\begin{equation}
\label{eq:test_of_adequacy}
\thinmuskip=0mu\medmuskip=0mu\thickmuskip=0mu
    \lVert\mathbf{t}+\lambda_0\mathbf{R}\widehat{\mathbf{f}}_0-\lambda_1\widehat{\mathbf{f}}_1\rVert^2 
    \ \geq \ 
    \min
    \left(
    \lVert\mathbf{t}+\lambda_0\mathbf{R}\widehat{\mathbf{f}}_0+\lambda_1\widehat{\mathbf{f}}_1\rVert^2, 
    \lVert\mathbf{t}-\lambda_0\mathbf{R}\widehat{\mathbf{f}}_0-\lambda_1\widehat{\mathbf{f}}_1\rVert^2,
    \lVert\mathbf{t}-\lambda_0\mathbf{R}\widehat{\mathbf{f}}_0+\lambda_1\widehat{\mathbf{f}}_1\rVert^2
    \right) \vspace{-0.2em}
\end{equation}
For the classic midpoint method, letting $\lambda_0 = |{\lambda_\text{mid}}_0|$ and $\lambda_1 = |{\lambda_\text{mid}}_1|$  gives effectively the same result as the cheirality check.
For example, \eqref{eq:test_of_adequacy} holds in Fig \ref{fig:cheirality_and_weighting}{\color{red}a} because the two points are closest when $\lambda_0 = -|{\lambda_\text{mid}}_0|$ and $\lambda_1 = -|{\lambda_\text{mid}}_1|$.

\vspace{-0.5em}
\subsection{Inverse Depth Weighted Midpoint}
\label{subsec:inverse_depth_weighted_midpoint}

The unweighted midpoint given by \eqref{eq:unweighted_midpoint} often entails disproportionate reprojection errors in the two images.
Fig. \ref{fig:cheirality_and_weighting}{\color{red}c} shows an example.
Notice that the ray with a smaller depth tends to yield a larger reprojection error.
To compensate this imbalance, we propose to use the inverse depth \cite{inverse_depth} as a weight:
\vspace{-0.2em}
\begin{equation}
\label{eq:weighted_midpoint}
    \mathbf{x}'_1 
    = \frac{\lambda_0^{-1}\left(\mathbf{t}+\lambda_0\mathbf{R}\widehat{\mathbf{f}}_0\right)+\lambda_1^{-1}\left(\lambda_1\widehat{\mathbf{f}}_1\right)}{\lambda_0^{-1}+\lambda_1^{-1}}
    \stackrel{\eqref{eq:lambdas2}}{=}\frac{\lVert\mathbf{q}\rVert}{\lVert\mathbf{q}\rVert+\lVert\mathbf{r}\rVert}\left(\mathbf{t}+\frac{\lVert\mathbf{r}\rVert}{\lVert\mathbf{p}\rVert}\left(\mathbf{R}\widehat{\mathbf{f}}_0+\widehat{\mathbf{f}}_1\right)\right).
\end{equation}

\newpage
\begin{figure}[t]
 \centering
 \includegraphics[width=\textwidth]{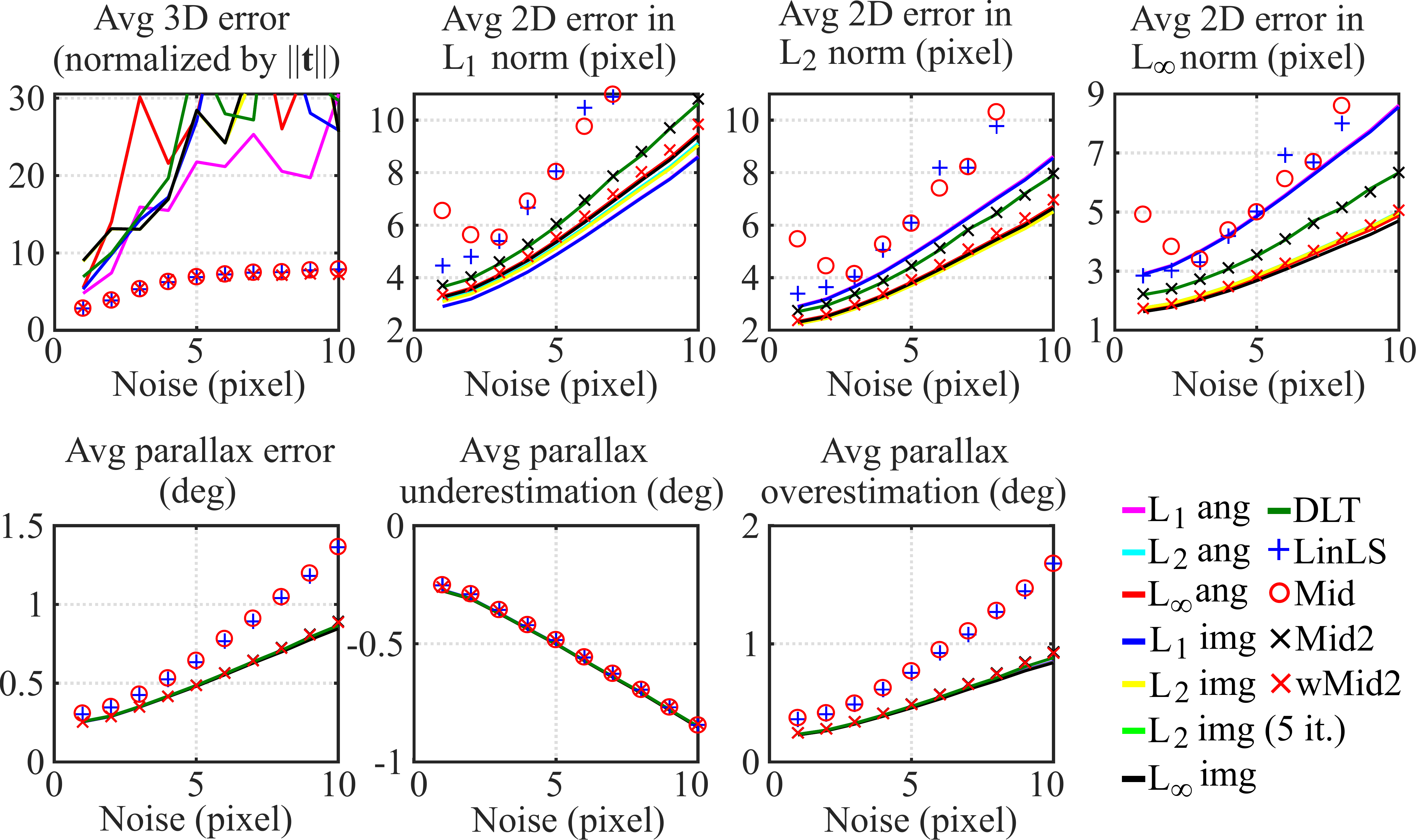}
\caption{
Triangulation accuracy over different noise levels in the image measurements.\hfill\break
\textbf{3D results (top 1$^\text{st}$ column):} 
LinLS, Mid, Mid2 and wMid2 perform almost equally and all significantly better than the rest. 
Among the optimal methods, $L_1$ ang performs best.\break
\textbf{2D results (top 2$^{\textbf{nd}}$--4$^{\textbf{th}}$ column):} 
LinLS and Mid perform worst across all norms.
Mid2 performs much better than those two, and wMid2 performs consistently better than Mid2.
Expectedly, optimal methods perform best in their respective error criterion. 
However, the differences among them are smaller in $L_1$ norm than in the other two norms.\break
\textbf{Parallax results (bottom row):}
LinLS and Mid perform worst, and the rest almost equally.
Looking at the under- and overestimation of the parallax separately, we notice that the low accuracy of LinLS and Mid is caused by their bias to overestimate the parallax on average.
}
\label{fig:stereo_errors_over_noise}
\vspace{-0.5em}
\end{figure}

\begin{figure}[ht]
 \centering
 \includegraphics[width=\textwidth]{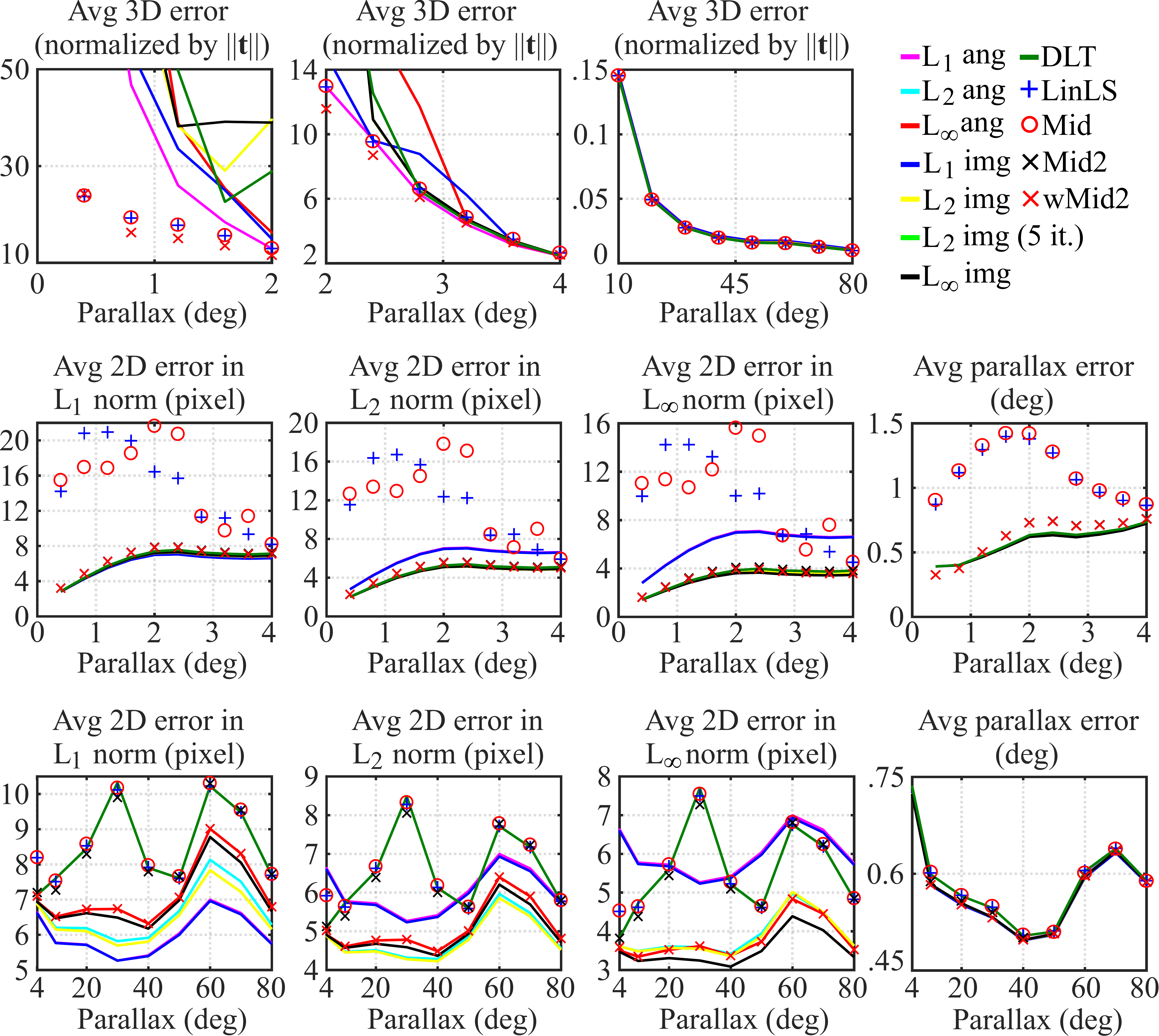}
\caption{
Triangulation accuracy over different parallax angles $\beta_\text{raw}$ \eqref{eq:raw_parallax}.\hfill\break
\textbf{3D results (top row):} 
For parallax under 2 deg, LinLS, Mid, Mid2 and wMid2 outperform the rest. 
For parallax between 2--4 deg, $L_1$ ang and the four aforementioned methods perform best.
For parallax over 4 deg, all methods perform almost equally.\break
\textbf{2D results for small parallax (mid 1$^\text{st}$--3$^\text{rd}$ column):} 
LinLS and Mid perform significantly worse than the rest in all norms.
Aside from those two, $L_1$ methods perform much worse than the rest in $L_2$ and $L_\infty$ norm, yet best in $L_1$ norm. 
The remaining methods perform similarly. \break
\textbf{2D results for large parallax (bottom 1$^\text{st}$--3$^\text{rd}$ column):} 
In all norms, LinLS, Mid, Mid2 and DLT perform consistently worse than wMid2, $L_2$ and $L_\infty$ methods.
The latter perform similarly and much better than $L_1$ methods in $L_2$ and $L_\infty$ norm, yet worse in $L_1$ norm.\hfill\break
\textbf{Parallax results (last column):}
For low-parallax points, LinLS and Mid perform worst (see Fig. \ref{fig:stereo_parallax_results_over_parallax} for more details). 
For high-parallax points, all methods perform equally well.
}
\label{fig:stereo_errors_over_parallax}
\vspace{-10em}
\end{figure}

\section{Evaluation Results}
\label{sec:results}
\vspace{-0.5em}
We evaluate the following methods:
Lee and Civera's $L_1$, $L_2$ and $L_\infty$ optimal angular methods (`$L_1$ ang', `$L_2$ ang', `$L_\infty$ ang') \cite{closed_form_optimal_triangulation_based_angular_errors}, Hartley and Sturm's $L_1$ and $L_2$ optimal methods (`$L_1$ img', `$L_2$ img') and linear methods (`DLT', `LinLS') \cite{hartley_triangulation, hartley_book}, Lindstrom's $L_2$ method with five iterations (`$L_2$ img (5 it.)') \cite{lindstrom}, N\'{i}ster's $L_\infty$ method (`$L_\infty$ img') \cite{nister_phd_thesis}, the classic midpoint method (`Mid') \cite{beardsley_midpoint1,beardsley_midpoint2, hartley_triangulation}, and our method without and with the weighting (`Mid2', `wMid2').
The evaluation was performed on synthetic datasets generated as follows:
A set of $8\times8$ point clouds of 5,000 points each are generated with a Gaussian radial distribution $\mathcal{N}(0,(d/4)^2)$ where $d$ is the distance from the world origin.
Each point cloud is centered at $[0,0,d]^\intercal$ for $d=2^n$ with $n=-1,0,...,+6$, and their image projections are perturbed by Gaussian noise $\mathcal{N}(0,\sigma^2)$ for $\sigma=1,2...,8$.
The size and the focal length of the images are $1,024^2$ pixels and $512$ pixel, respectively.
We have four configurations for the camera poses: (1) `orbital' - two cameras at $[\pm0.5,0,0]^\intercal$ pointing at the point cloud center, (2) `lateral' - two cameras at $[\pm0.5,0,0]^\intercal$ pointing at $[0,0,\infty]^\intercal$, (3) `forward' - two cameras at $[0,0,\pm0.5]^\intercal$ pointing at the point cloud center, and (4) `diagonal' - two cameras at $\pm[\sqrt{3}/6,\sqrt{3}/6,\sqrt{3}/6]^\intercal$ pointing at $[0,0,\infty]^\intercal$.
The poses are slightly perturbed with uniform noise $\mathcal{U}(0, 0.01)$.
In total, the datasets provide over a million unique triangulation problems.
\newpage

\vspace*{\fill}
We aggregate the results in Fig. \ref{fig:stereo_errors_over_noise} and \ref{fig:stereo_errors_over_parallax}.
Our observations agree with \cite{hartley_triangulation} in that:
\vspace{-0.3em}
\begin{enumerate}\itemsep0em
    \item 
    Generally, greater noise and lower parallax lead to larger 3D errors. 
    All methods yield almost equally low 3D errors for high-parallax points (> 4 deg). \vspace{-0.3em}
    \item 
    2D and 3D errors are not well correlated.
    For example, LinLS and Mid perform best in 3D, but worst in 2D.
\end{enumerate}

\newpage
\begin{figure}[t]
 \centering
 \includegraphics[width=\textwidth]{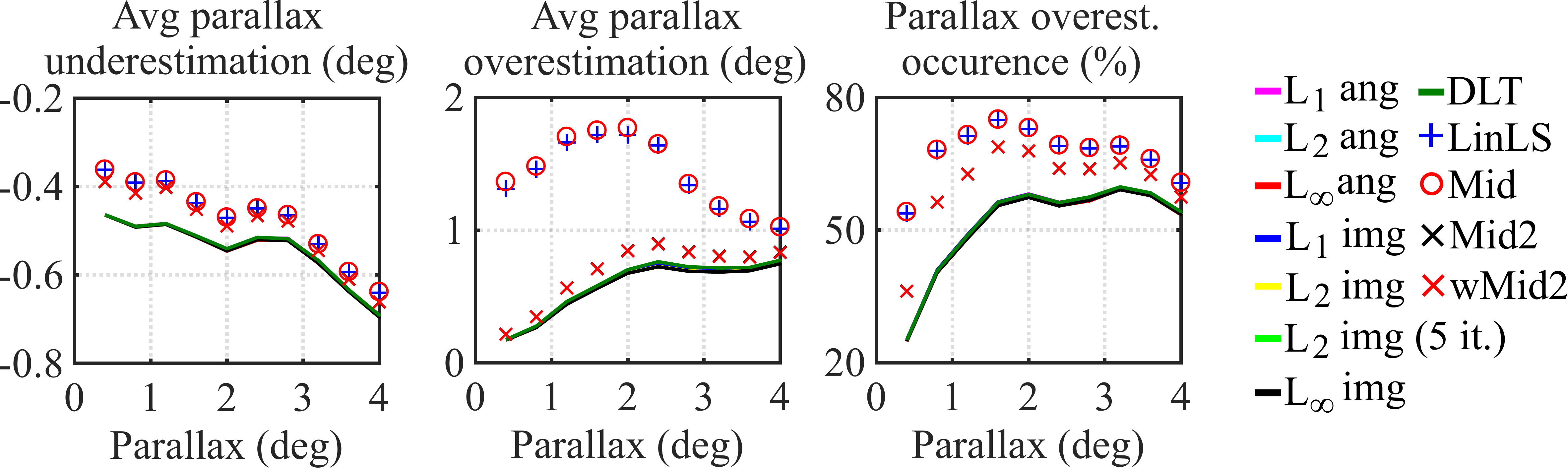}
\caption{
Parallax accuracy for low-parallax points. 
LinLS and Mid are relatively more biased to overestimate the small parallax angles, both in magnitude and frequency.
The same goes for Mid2 and wMid2, but to a lesser extent.
}
\label{fig:stereo_parallax_results_over_parallax}
\end{figure}

\begin{wrapfigure}{R}{0.45\textwidth}
 \centering
 \vspace{-1em}
 \includegraphics[width=0.4\textwidth]{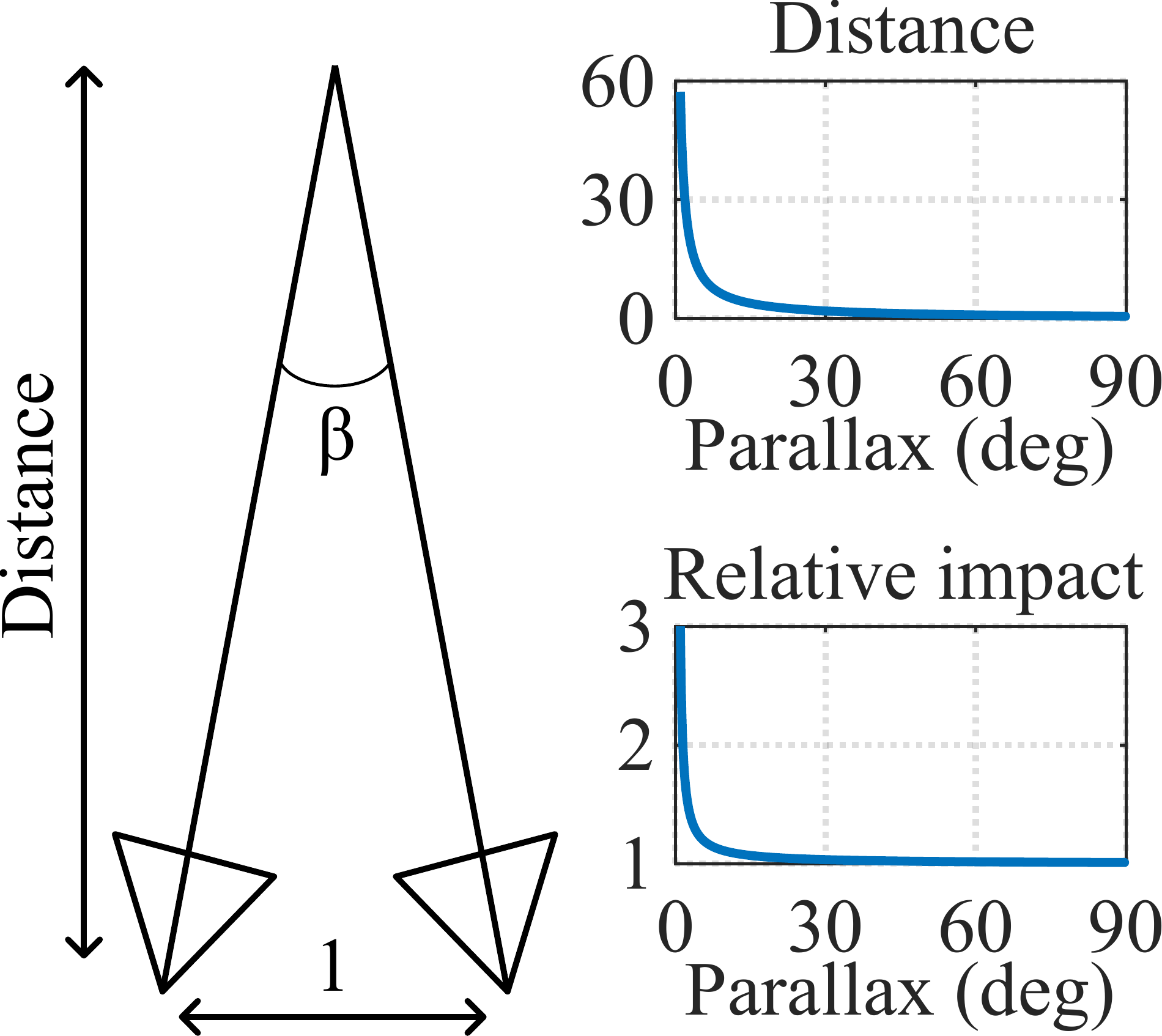}
\caption{
\textbf{Top:} In this 2D example, the distance is proportional to $\cot{(\beta/2)}$.
\textbf{Bottom:} Relative impact is defined as $|{f(\beta)}/{g(\beta)}|$ where $f(\beta)$ and $g(\beta)$ are the distance errors when the parallax is under- and overestimated by 0.5 deg, respectively.
Underestimating the parallax angle yields a larger distance error than overestimating it by the same degree, and especially more so for smaller parallax.
}
\vspace{-15em}
\label{fig:parallax}
\end{wrapfigure}

Additionally, we report the following findings:
\vspace{-0.3em}
\begin{enumerate}\itemsep0em
    \item 
    It is difficult to tell which method is the best in terms of 2D accuracy. 
    For example, $L_1$ methods yield the lowest 2D errors in $L_1$ norm, but relatively larger errors in $L_2$ and $L_\infty$ norm.
    It is not obvious which norm is more important.
    That said, some methods can still perform consistently better than others;
    wMid2, $L_2$ and $L_\infty$ methods consistently outperform LinLS, Mid, Mid2 and DLT in all 2D error criteria.
    \vspace{-0.3em}
    \item
    As shown in Fig. \ref{fig:stereo_parallax_results_over_parallax}, LinLS and Mid are clearly more biased to overestimate the small parallax angles (< 4 deg). 
    This explains their relatively low 3D errors at low parallax. 
    Fig. \ref{fig:parallax} illustrates a simplified example of this effect.
    \vspace{-0.3em}
    \item 
    Our methods (Mid2 and wMid2) achieve the best overall accuracy in 3D + parallax. \vspace{-0.3em}
\end{enumerate}

The last finding can be solely attributed to the low-parallax points, for which our methods show similar 3D accuracy to that of Mid, but much better parallax accuracy.
The latter can be explained by the fact that Mid2 always yields larger depths than Mid (as discussed in Section \ref{subsec:alternative_midpoint_method}), which in effect lowers the estimated parallax angle on average (as shown in Fig. \ref{fig:stereo_parallax_results_over_parallax}).
Since Mid tends to overestimate small parallax angles, this works to our advantage at low parallax.

\noindent\begin{tabular}{@{}p{0.52\textwidth}}
\hspace{1em} 
The first plot in Fig. \ref{fig:stereo_parallax_results_over_parallax} shows that Mid, LinLS and our two methods underestimate small parallax angles slightly less than the rest.
It is no coincidence that precisely these four methods achieve the best 3D accuracy; 
Fig. \ref{fig:parallax} suggests that underestimating a small parallax angle has a severe impact on 3D accuracy.
At low parallax, it seems that our methods hit the sweet spot by (1) underestimating less than the optimal methods and DLT (thus achieving lower 3D errors) and (2) overestimating less than Mid and LinLS (thus achieving lower parallax errors). 
\vspace{-5em}
\end{tabular}

\newpage
We suspect that some of the large 2D errors of Mid and LinLS at low parallax are related to their large parallax overestimation (see the second row of Fig. \ref{fig:stereo_errors_over_parallax}).
Large reprojection errors mean that the rays were corrected by a large amount, and pivoting two almost parallel rays (i.e., rays that correspond to a low-parallax point) will most likely increase the angle between them.
This link between the 2D and the parallax error could partially explain why our methods yield smaller 2D errors than Mid and LinLS for low-parallax points.

We also found that the 2D errors of wMid2 resemble those of $L_\infty$ methods (see the bottom row of Fig. \ref{fig:stereo_errors_over_parallax}).
Note that the $L_\infty$ optimal solutions yield the equal reprojection errors in the two views \cite{nister_phd_thesis, closed_form_optimal_triangulation_based_angular_errors}.  
This suggests that our inverse depth weighting (Section \ref{subsec:inverse_depth_weighted_midpoint}) not only reduces the overall 2D errors, but also balances the reprojection errors in the two images.

\begin{figure}[t]
 \centering
 \includegraphics[width=\textwidth]{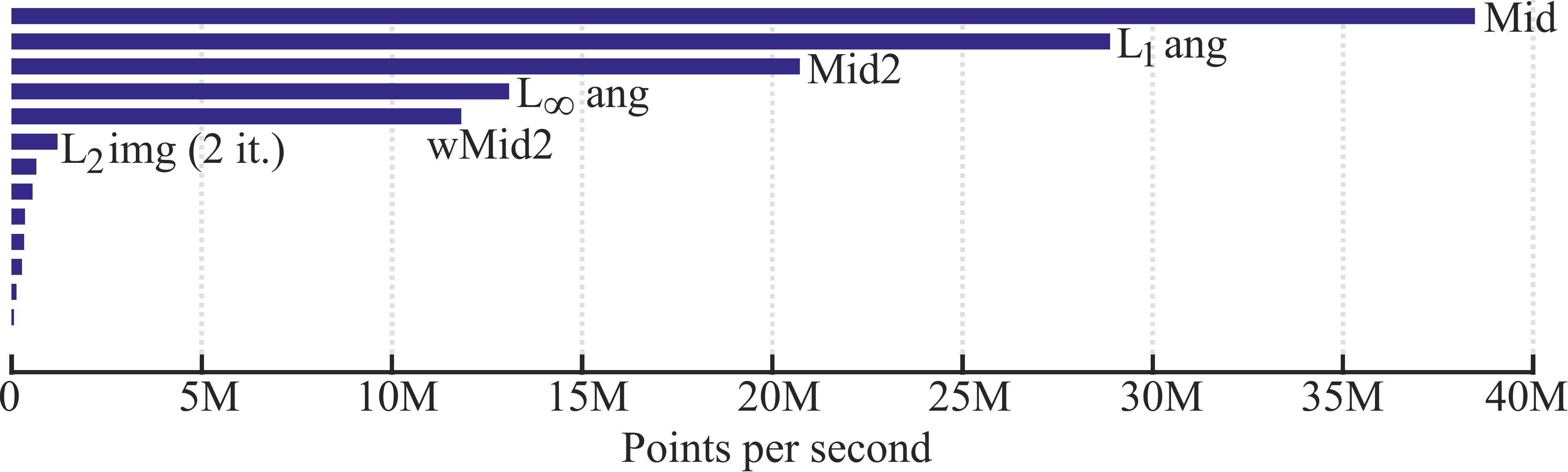}
\caption{
Speed of computing a 3D point. In descending order: Mid (38M), $L_1$ ang (29M), Mid2 (21M), $L_\infty$ ang (13M), wMid2 (12M), $L_2$ img 2 it. (1.2M), $L_2$ ang (650K), $L_2$ img 5 it. (550K), LinLS (350K), DLT (330K), $L_\infty$ img (270K), $L_2$ img (120K), $L_1$ img (65K).
}
\label{fig:timings}
\end{figure}

Fig. \ref{fig:timings} compares the speed of each method. 
We included the test of the adequacy (Section \ref{subsec:cheirality}) in both our methods.
All methods were implemented in C++ using the Eigen library \cite{eigen_library}, compiled using GCC \cite{gcc} with -O3 level optimization, and run on a laptop CPU (Intel i7-4810MQ, 2.8 GHz).
Although wMid2 is almost two times slower than Mid2 and three times than Mid, it is still at least ten times faster than the state-of-the-art method \cite{lindstrom}.
Notice that even though \eqref{eq:lambdas2} and \eqref{eq:lambdas_mid} are similar, Mid is still almost twice faster than Mid2.
This happens for two reasons: 
First, we avoid computing the square root in \eqref{eq:lambdas_mid} by multiplying the numerator and denominator by $\lVert\mathbf{p}\rVert$ and get ${\lambda_\text{mid}}_0 = \frac{\mathbf{p}^\intercal\mathbf{r}}{\mathbf{p}^\intercal\mathbf{p}}$ and ${\lambda_\text{mid}}_1 = \frac{\mathbf{p}^\intercal\mathbf{q}}{\mathbf{p}^\intercal\mathbf{p}}$.
Second, the test of adequacy described in Section \ref{subsec:cheirality} takes longer than the standard cheirality check.
Nevertheless, given that the computational cost of triangulation is relatively small compared to other operations (e.g., point matching, pose estimation and structure refinement) \cite{hartley_triangulation}, our methods provide an excellent trade-off between speed and accuracy.

\section{Discussions}
\label{sec:discussion}
\vspace{-0.5em}
\subsection{On Optimality}
A reasonable doubt is that the proposed (weighted) midpoint method may be in fact optimal in some error criterion. 
If we consider algebraic errors, it is obvious that our midpoint minimizes $\lVert\mathbf{x}'_1-\mathbf{g}(\mathbf{R},\mathbf{t},\mathbf{f}_0, \mathbf{f}_1)\rVert$ where $\mathbf{g}(\cdot)$ is the vector-valued function given by the right-hand side of \eqref{eq:unweighted_midpoint} or \eqref{eq:weighted_midpoint}.
However, there are infinitely many other error functions for which the global minimum is $\mathbf{g}(\mathbf{R},\mathbf{t},\mathbf{f}_0, \mathbf{f}_1)$, and it is hard to tell which one is geometrically meaningful (at least, we could not find it or disprove its existence).

Another reasonable doubt is that, among the GWM methods, there could be a better one than ours, since we have not proved the optimality of the proposed weighted midpoint.
A more fundamental question is then: 
What accuracy measure should we choose to define the optimality, knowing that there is a discrepancy between different types of errors, e.g., image/angular reprojection errors in different norms, 3D and parallax errors?
At least for our method, it seems that the weighting affects mostly the 2D errors, so there might be a certain weighting scheme that guarantees 2D optimality without compromising 3D and parallax accuracy.
A more elaborate error analysis and the comparison of different weightings within the GWM framework remain for future work.
For more discussion of the optimality in geometric vision, we refer to \cite{optimal_algorithms_hartley} and Appendix 3 of \cite{hartley_book}.

\vspace{-0.5em}
\subsection{On Practical Implications}
In many SfM pipelines, two-view triangulation is used to initialize the 3D map points prior to the bundle adjustment \cite{snavely2006photo,sfm_revisited,nousias2019large}.
Since the points with small parallax angles are associated with large 3D uncertainty, they are usually discarded.
This strategy is viable if there are enough correspondences with large parallax angles.
However, it is not ideal, as (1) low-parallax points can be useful for camera orientation estimation \cite{civera2008interacting,pirchheim2013handling} and (2) if the goal itself is to reconstruct the scene with large depths compared to the baseline.
For problems such as reconstruction from small-baseline (or accidental) motions \cite{delon2007small,accidental_motion}, small parallax angles are quite common, so our method could be relevant.
Extending our method to multiple views for reconstructing low-parallax scenes would be an interesting future direction.

\vspace{-0.5em}
\section{Conclusions}
\label{sec:conclusion}
Triangulation from two views with known calibration and pose is an age-old problem in computer vision.
Existing methods formulate the problem as the minimization of some cost function, most commonly reprojection errors.
In this paper, we asked ourselves if this is really the best approach.
To this end, we proposed a novel variant of the classic midpoint method that does not minimize geometric or algebraic errors.

We found that all the existing methods we evaluated perform poorly at low parallax, producing large errors in either 2D, 3D or parallax. 
On the other hand, our midpoint method achieves very good overall accuracy.
We also showed that incorporating the inverse depth weighting can further reduce the 2D errors.
Although our method is not theoretically optimal, it provides, with speed and simplicity, a superior balance of 2D, 3D and parallax accuracy in practice.

\section*{Acknowledgement}
We thank Christopher Mei at Microsoft for valuable discussions.
This work was partially supported by the Spanish government (project PGC2018-096367-B-I00) and the Arag{\'{o}}n regional government (Grupo DGA-T45{\_}17R/FSE).

\section*{Appendix}
In the following, we derive \eqref{eq:lambdas_mid}.
In literature, the formula has been used many times without derivation \cite{kanazawa, lindstrom, closed_form_optimal_triangulation_based_angular_errors}.
For the sake of completeness, we present the full derivations here.
In doing so, we will use the following properties of the dot product and the cross product operations:
\begin{gather}
    \widehat{\mathbf{a}}\times (\widehat{\mathbf{a}}\times\mathbf{b})=\widehat{\mathbf{a}} (\widehat{\mathbf{a}}\cdot\mathbf{b})-\mathbf{b}.  \label{eq:ax(axb)}\\
    (\widehat{\mathbf{a}}\times\mathbf{b})\cdot(\widehat{\mathbf{a}}\times\mathbf{c})=\mathbf{b}\cdot\mathbf{c}-(\widehat{\mathbf{a}}\cdot\mathbf{b})(\widehat{\mathbf{a}}\cdot\mathbf{c}). \label{eq:(axb).(axc)}\\
    ({\mathbf{a}}\times{\mathbf{b}})\times({\mathbf{a}}\times{\mathbf{c}})=\left(\mathbf{a}\cdot(\mathbf{b}\times\mathbf{c})\right)\mathbf{a}. \label{eq:(axb)x(axc)}
\end{gather}
Next, we introduce the following lemma:
\begin{lemma}[The Closest Pair of Points on Two Skew Lines]
\label{lemma:closest_points}
Consider two skew lines $\mathbf{L}_0(s_0) = \mathbf{c}_0+s_0{\mathbf{m}}_0$ and $\mathbf{L}_1(s_1) = \mathbf{c}_1+s_1{\mathbf{m}}_1$ in 3D space. 
Let $\mathbf{t}=\mathbf{c}_0-\mathbf{c}_1$ and $\left(\mathbf{r}_0, \mathbf{r}_1\right)$ be the two points on each line that form the closest pair.
Then, 
\begin{equation}
    \label{eq:lemma_closest_ponits1}
    \mathbf{r}_0 = \mathbf{c}_0+\frac{\left(\widehat{\mathbf{m}}_0\times\widehat{\mathbf{m}}_1\right)\cdot\left(\widehat{\mathbf{m}}_1\times\mathbf{t}\right)}{\lVert\widehat{\mathbf{m}}_0\times\widehat{\mathbf{m}}_1\rVert^2}\widehat{\mathbf{m}}_0
\end{equation}
and 
\begin{equation}
    \label{eq:lemma_closest_ponits2}
    \mathbf{r}_1 = \mathbf{c}_1+\frac{\left(\widehat{\mathbf{m}}_0\times\widehat{\mathbf{m}}_1\right)\cdot\left(\widehat{\mathbf{m}}_0\times\mathbf{t}\right)}{\lVert\widehat{\mathbf{m}}_0\times\widehat{\mathbf{m}}_1\rVert^2}\widehat{\mathbf{m}}_1.
\end{equation}
\end{lemma}
\noindent\textbf{Proof. }  
In geometry, it is a well-known fact that the closest pair of points on two skew lines lie on the common perpendicular to both lines.
In other words, the vector $\mathbf{r}_0-\mathbf{r}_1$ is perpendicular to both $\mathbf{L}_0$ and $\mathbf{L}_1$.
Therefore, for some scalar $\tau$, 
\begin{equation}
\label{eq:lemma_closest_points_proof1}
    \mathbf{r}_0-\mathbf{r}_1 = \tau \left(\widehat{\mathbf{m}}_0\times\widehat{\mathbf{m}}_1\right).
\end{equation}
Since point $\mathbf{r}_0$ and $\mathbf{r}_1$ are respectively located along $\mathbf{L}_0$ and $\mathbf{L}_1$, we can write 
\begin{equation}
    \label{eq:lemma_closest_points_proof1.1}
    \mathbf{r}_0=\mathbf{c}_0+\lambda_0\widehat{\mathbf{m}}_0 \quad  \text{and} \quad  \mathbf{r}_1=\mathbf{c}_1+\lambda_1\widehat{\mathbf{m}}_1.
\end{equation}
for some scalar $\lambda_0$ and $\lambda_1$. 
Then, \eqref{eq:lemma_closest_points_proof1} becomes
\begin{equation}
\label{eq:lemma_closest_points_proof2}
    \mathbf{t}+\lambda_0\widehat{\mathbf{m}}_0-\lambda_1\widehat{\mathbf{m}}_1=\tau\mathbf{n},
\end{equation}
where $\mathbf{n}=\widehat{\mathbf{m}}_0\times\widehat{\mathbf{m}}_1$.
This makes a system of three equations (in each coordinate x, y and z) with three unknowns  $\lambda_0$, $\lambda_1$, and $\tau$.
Removing $\tau$ from the equations leads to
\begin{equation}
\label{eq:lemma_closest_points_proof3}
    \frac{t_x+\lambda_0m_{0x}-\lambda_1m_{1x}}{n_x}
    =
    \frac{t_y+\lambda_0m_{0y}-\lambda_1m_{1y}}{n_y}
\end{equation}
and
\begin{equation}
\label{eq:lemma_closest_points_proof4}
    \frac{t_y+\lambda_0m_{0y}-\lambda_1m_{1y}}{n_y}
    =
    \frac{t_z+\lambda_0m_{0z}-\lambda_1m_{1z}}{n_z}.
\end{equation}
Note that $\mathbf{t}=[t_x, t_y, t_z]^\intercal$, $\mathbf{n}=[n_x, n_y, n_z]^\intercal$, $\widehat{\mathbf{m}}_0 = [m_{0x}, m_{0y}, m_{0z}]^\intercal$ and $\widehat{\mathbf{m}}_1 = [m_{1x}, m_{1y}, m_{1z}]^\intercal$.
From \eqref{eq:lemma_closest_points_proof3} and \eqref{eq:lemma_closest_points_proof4}, we get
\begin{equation}
\label{eq:lemma_closest_points_proof5}
    \lambda_0 = \frac{\lambda_1\left(m_{1x}n_y-m_{1y}n_x\right)+t_yn_x-t_xn_y}{m_{0x}n_y-m_{0y}n_x}
\end{equation}
and
\begin{equation}
\label{eq:lemma_closest_points_proof6}
    \lambda_0 = \frac{\lambda_1\left(m_{1y}n_z-m_{1z}n_y\right)+t_zn_y-t_yn_z}{m_{0y}n_z-m_{0z}n_y}.
\end{equation}
Equating the right-hand sides of \eqref{eq:lemma_closest_points_proof5} and \eqref{eq:lemma_closest_points_proof6} leads to
\begin{equation}
\label{eq:lemma_closest_points_proof6.1}
    \lambda_1 = \frac{A-B}{C-D},
\end{equation}
where
\begin{align}
    A&= (t_zn_y-t_yn_z)(m_{0x}n_y-m_{0y}n_x)\nonumber\\
    B&= (t_yn_x-t_xn_y)(m_{0y}n_z-m_{0z}n_y)\nonumber\\
    C&= (m_{1x}n_y-m_{1y}n_x)(m_{0y}n_z-m_{0z}n_y) \nonumber\\
    &=\left(\widehat{\mathbf{m}}_1\times\mathbf{n}\right)_z\left(\widehat{\mathbf{m}}_0\times\mathbf{n}\right)_x,\nonumber\\
    D&= (m_{1y}n_z-m_{1z}n_y)(m_{0x}n_y-m_{0y}n_x)\nonumber\\
    &= \left(\widehat{\mathbf{m}}_1\times\mathbf{n}\right)_x\left(\widehat{\mathbf{m}}_0\times\mathbf{n}\right)_z.\nonumber
\end{align}
We can rearrange $A-B$ into
\begin{equation}
\label{eq:lemma_closest_points_proof7}
    A-B = n_y\mathbf{t}\cdot
    \begin{pmatrix}
    m_{0y}n_z-m_{0z}n_y \\ m_{0z}n_x-m_{0x}n_z \\
    m_{0x}n_y-m_{0y}n_x
    \end{pmatrix}.
\end{equation}
The latter term in the dot product of \eqref{eq:lemma_closest_points_proof7} is equal to $\widehat{\mathbf{m}}_0\times\mathbf{n}$. 
Thus,
\begin{align}
    A-B &= n_y\mathbf{t}\cdot\left(\widehat{\mathbf{m}}_0\times\left(\widehat{\mathbf{m}}_0\times\widehat{\mathbf{m}}_1\right)\right)\nonumber\\
    &\stackrel{\eqref{eq:ax(axb)}}{=}n_y\mathbf{t}\cdot\left(\widehat{\mathbf{m}}_0\left(\widehat{\mathbf{m}}_0\cdot\widehat{\mathbf{m}}_1\right)-\widehat{\mathbf{m}}_1\right)\nonumber\\
    &=n_y\left(\left(\widehat{\mathbf{m}}_0\cdot\widehat{\mathbf{m}}_1\right)\left(\widehat{\mathbf{m}}_0\cdot\mathbf{t}\right)-\widehat{\mathbf{m}}_1\cdot\mathbf{t}\right)\nonumber\\
    &\stackrel{\eqref{eq:(axb).(axc)}}{=}-n_y\left(\widehat{\mathbf{m}}_0\times\widehat{\mathbf{m}}_1\right)\cdot\left(\widehat{\mathbf{m}}_0\times\mathbf{t}\right).\label{eq:lemma_closest_points_proof8}
\end{align}
We can rearrange $C-D$ into
\begin{align}
    C-D&=\left(\widehat{\mathbf{m}}_1\times\mathbf{n}\right)_z\left(\widehat{\mathbf{m}}_0\times\mathbf{n}\right)_x-\left(\widehat{\mathbf{m}}_1\times\mathbf{n}\right)_x\left(\widehat{\mathbf{m}}_0\times\mathbf{n}\right)_z \nonumber\\
    &=\left(\left(\widehat{\mathbf{m}}_1\times\mathbf{n}\right)\times\left(\widehat{\mathbf{m}}_0\times\mathbf{n}\right)\right)_y \nonumber\\
    &\stackrel{\eqref{eq:(axb)x(axc)}}{=} \left(\left(\mathbf{n}\cdot\left(\widehat{\mathbf{m}}_1\times\widehat{\mathbf{m}}_0\right)\right)\mathbf{n}\right)_y \nonumber\\
    &=\left(-\lVert\widehat{\mathbf{m}}_0\times\widehat{\mathbf{m}}_1\rVert^2\mathbf{n}\right)_y \nonumber\\
    &=-\lVert\widehat{\mathbf{m}}_0\times\widehat{\mathbf{m}}_1\rVert^2 n_y. \label{eq:lemma_closest_points_proof9}
    \end{align}
Substituting \eqref{eq:lemma_closest_points_proof8} and \eqref{eq:lemma_closest_points_proof9} into \eqref{eq:lemma_closest_points_proof6.1} gives
\begin{equation}
    \label{eq:lemma_closest_points_proof10}
    \lambda_1 = \frac{\left(\widehat{\mathbf{m}}_0\times\widehat{\mathbf{m}}_1\right)\cdot\left(\widehat{\mathbf{m}}_0\times\mathbf{t}\right)}{\lVert\widehat{\mathbf{m}}_0\times\widehat{\mathbf{m}}_1\rVert^2}.
\end{equation}
Finally, substituting \eqref{eq:lemma_closest_points_proof10} into \eqref{eq:lemma_closest_points_proof1.1} leads to \eqref{eq:lemma_closest_ponits2}.
Equation \eqref{eq:lemma_closest_ponits1} is derived analogously. \QEDA

\vspace{1em}

By substituting $\mathbf{Rf}_0$ and $\mathbf{f}_1$ into $\mathbf{m}_0$ and $\mathbf{m}_1$, we can use lemma \ref{lemma:closest_points} to obtain ${\lambda_\text{mid}}_0$ and ${\lambda_\text{mid}}_1$ in \eqref{eq:lambdas_mid}.

\begin{equation}
\label{eq:lemma_closest_points_proof11}
    {\lambda_\text{mid}}_0 = \frac{(\mathbf{R}\widehat{\mathbf{f}}_0\times\widehat{\mathbf{f}}_1)\cdot(\widehat{\mathbf{f}}_1\times\mathbf{t})}{\lVert\mathbf{R}\widehat{\mathbf{f}}_0\times\widehat{\mathbf{f}}_1\rVert^2},
    \quad
     {\lambda_\text{mid}}_1 = \frac{(\mathbf{R}\widehat{\mathbf{f}}_0\times\widehat{\mathbf{f}}_1)\cdot(\mathbf{R}\widehat{\mathbf{f}}_0\times\mathbf{t})}{\lVert\mathbf{R}\widehat{\mathbf{f}}_0\times\widehat{\mathbf{f}}_1\rVert^2}.
\end{equation}
Letting $\mathbf{p} = \mathbf{R}\widehat{\mathbf{f}}_0\times\widehat{\mathbf{f}}_1$, $\mathbf{q} = \mathbf{R}\widehat{\mathbf{f}}_0\times\mathbf{t}$ and $\mathbf{r} =  \widehat{\mathbf{f}}_1\times\mathbf{t}$ leads to \eqref{eq:lambdas_mid}.

\end{document}